\documentclass{bmvc2k}

\usepackage{graphicx}
\usepackage{microtype}
\usepackage{amsmath}
\usepackage{amssymb}
\usepackage{algorithm}
\usepackage{color}

\usepackage{multicol}
\usepackage{multirow}
\usepackage{dsfont}
\usepackage{mathtools}
\usepackage{enumitem}
\usepackage{colortbl}
\usepackage{arydshln}
\definecolor{headercolor}{gray}{0.93}
\newcolumntype{a}{>{\columncolor{headercolor}}c}
\usepackage{booktabs,multirow,multicol}
\newcommand{\revise}[1]{\textcolor{black}{{#1}}}

\title{Leveraging Human Selective Attention for Medical Image Analysis with Limited Training Data}

\addauthor{Yifei Huang}{hyf@iis.u-tokyo.ac.jp}{1}
\addauthor{Xiaoxiao Li}{xiaoxiao.li@ece.ubc.ca}{2}
\addauthor{Lijin Yang}{yang-lj@iis.u-tokyo.ac.jp}{1}
\addauthor{Lin Gu}{lin.gu@riken.jp}{3}
\addauthor{Yingying Zhu}{yingying.zhu@uta.edu}{4}
\addauthor{Hirofumi Seo, Qiuming Meng}{hseotky@gmail.com, mengqiuming1993@gmail.com}{1}
\addauthor{Tatsuya Harada, Yoichi Sato}{harada@mi.t.u-tokyo.ac.jp, ysato@iis.u-tokyo.ac.jp}{1}

\addinstitution{
 The University of Tokyo\\ Japan
}
\addinstitution{
The University of British Columbia \\ Canada
}
\addinstitution{
RIKEN, AIP \\ Japan
}
\addinstitution{
University of Texas Arlington \\ USA
}

\runninghead{Huang et al.}{Leveraging Human Selective Attention}

\def\etal{\emph{et al}\bmvaOneDot}

\begin{document}

\maketitle

\begin{abstract}
Human gaze is a cost-efficient physiological data that reveals human underlying attentional patterns. The selective attention mechanism helps the cognition system focus on task-relevant visual clues by ignoring the presence of distractors. Thanks to this ability, human beings can efficiently learn from a very limited number of training samples. Inspired by this mechanism, we aim to leverage gaze for medical image analysis tasks with small training data. Our proposed framework includes a backbone encoder and a Selective Attention Network (SAN) that simulates the underlying attention.
The SAN implicitly encodes information such as suspicious regions that is relevant to the medical diagnose tasks by estimating the actual human gaze. Then we design a novel Auxiliary Attention Block (AAB) to allow information from SAN to be utilized by the backbone encoder to focus on selective areas. Specifically, this block uses a modified version of a multi-head attention layer to simulate the human visual search procedure. Note that the SAN and AAB can be plugged into different backbones, and the framework can be used for multiple medical image analysis tasks when equipped with task-specific heads. Our method is demonstrated to achieve superior performance on both 3D tumor segmentation and 2D chest X-ray classification tasks. We also show that the estimated gaze probability map of the SAN is consistent with an actual gaze fixation map obtained by board-certified doctors.
\let\thefootnote\relax\footnotetext{\hspace{-1.9em}Corresponding author: Lin Gu}
\end{abstract}

\section{Introduction}
\label{sec:intro}

Human gaze behavior is ecologically linked to the spatial underlying attention~\cite{posner1990attention,Zhang2020GazeReview}. With numerous objects and rich information in view, human learns to develop a selective attentional mechanism that instinctively moves the fovea centralis to process the important task-relevant visual clues. 
This strategy is crucial to protect the brain, which has limited sensory and cognitive capacities, from being bombarded with an excess of information~\cite{broadbent2013perception}. 
“Without selective interest, experience is an utter chaos”~\cite{James1890}. Thanks to the selective attentive mechanisms, humans can efficiently learn from experience with limited samples. 

The selective attentional mechanism of the human gaze has been exploited to provide additional cues to improve the performance of multiple tasks, such as action recognition~\cite{huang2020mutual,li2021eye,min2021WACV}, object referring~\cite{vasudevan2018object}, and image captioning~\cite{he2019human}. A very recent work~\cite{liu2021goal} demonstrates that the estimation of gaze could effectively improve the classification even without directly accessing the target data (zero-shot learning). 

Inspired by the existing success, we explore the possibility of using human gaze supervised attention model to promote medical image analysis tasks under limited training data. We propose a novel framework that exploits the selective attention revealed by human gaze to allow efficient training with limited data. As illustrated in Fig.~\ref{fig:model}, we simulate the human's selective attention with a Selective Attention Network (SAN, bounded in blue), which is trained with actual human gaze collected from medical experts when they were performing the screening. We design a plug-and-play module that enables the information from SAN to be leveraged for directly enhancing the backbone performance on multiple tasks. \revise{This is achieved by a novel Auxiliary Attention Block (AAB) that integrates the information from SAN into the backbone. Noting the scaled dot-product attention in transformer~\cite{vaswani2017attention} is very similar to the attention exhibited by human gaze, we design the AAB based on a transformer architecture}. This block uses features from the backbone encoder as the \textit{key} vectors and \textit{value} vectors, and the representation from SAN as the \textit{query} vectors. For each query vector, the transformer computes its relation with all locations of the key vector to generate an attention map, which is very similar to the process that human medical experts' search suspicious regions on the image based on their spatial underlying attention. As a result, the embedded attention refines the features from the backbone encoder, making it better focus on important regions when conducting segmentation or classification tasks.

To evaluate the proposed framework, we conduct experiments on two public datasets. The first experiment is brain tumor segmentation on the BraTS 2020 dataset~\cite{bakas2018identifying}. We show that our method can achieve superior performance when compared to other state-of-the-art methods when using limited training data. We also collect gaze data of board-certified doctors for the BraTS 2020 dataset which could further contribute to the community. For a more complete study of our framework, we also experiment with chest X-ray classification tasks on the MIMIC-CXR-gaze dataset~\cite{karargyris2021creation}. Experiments demonstrate that our framework can exploit gaze information to benefit multiple backbone networks with only limited training data. The experiments with two kinds of tasks show the generalization ability of our framework.

Our contributions are summarized as follows:\vspace{-0.2cm}
\begin{itemize}
    \item We propose a novel framework that leverages human gaze to enhance the medical image tasks of classification and segmentation on both 2D and 3D data. This could promote medical AI's real-world clinical applications where labels are hard to collect. \vspace{-0.2cm}
    \item We collect new gaze data of board-certified doctors is for the BraTS 2020 dataset. \vspace{-0.2cm}
    \item  Experiments on the BraTS 2020 dataset~\cite{bakas2018identifying} and the MIMIC-CXR-gaze~\cite{karargyris2021creation} dataset demonstrate that our framework can significantly improve the performance of the backbone network on limited training data, with the help of human gaze in training.\vspace{-0.2cm}
\end{itemize}

\section{Related Works}
\textbf{Gaze assisted deep learning}
Since gaze is a cost-efficient physiological data that could be relatively easy to collect, there is a rising trend of collecting and utilizing the gaze data~\cite{li2021eye,karargyris2021creation,huang2020ego} to assist deep learning.
Noticing gaze are directly relevant to human attention or intention from the first-person viewpoint~\cite{li2013learning,huang2018predicting}, early researches mainly leverage gaze for first-person action recognition~\cite{fathi2012learning,huang2020mutual,li2015delving}. This is usually done by multi-task learning~\cite{kapidis2019multitask} or using gaze as soft attention maps~\cite{li2021eye}. However, the nice performance 
of these action classification methods could not be replicated on classification tasks such as chest X-ray classification~\cite{karargyris2021creation}. More recently,  human gaze are used to improve model performance in other tasks such as object detection~\cite{vasudevan2018object}, knowledge distillation~\cite{ArijitMICCAI19} and image captioning~\cite{he2019human}. A very recent work leveraged human gaze to improve image classification when no training sample of target class is given~\cite{liu2021goal}. However, this method relies on additional attribute information, and is not flexible for other tasks such as segmentation. In this work, we propose a novel framework that explores the use of gaze information to enhance multiple medical image analysis tasks including brain tumor segmentation and chest X-ray classification with limited data.

\textbf{Brain tumor segmentation}
Tumor segmentation is one of the most challenging task in medical imaging, because lesions are relative small compared to the whole image and may have various deformations, sizes or subtypes.  
DeepMedic~\cite{kamnitsas2016deepmedic} is a pioneering work showing superior performance for brain tumor segmentation using 3D patches with 3D CNN. Following DeepMedic, Casamitjana \textit{et al.}~\cite{casamitjana20163d} proposed a 3D CNN to take the entire 3D volume for prediction. 
Brain tumor segmentation performances on BraTS challenges~\cite{bakas2017advancing} are significantly boosted with 3D UNets~\cite{ronneberger2015u,cciccek20163d,chen2018s3d,kayalibay2017cnn,isensee2017brain}. 
3D CNN typically contains much more parameters than 2D CNN, thus its performance is constrained by the limited labeled training samples in the medical imaging domain. \revise{Although there are existing works directly using gaze data for brain tumor segmentation \cite{stember2019eye,stember2020integrating}, how to benefit model design on general medical tasks from gaze mechanism is under-explored.}

\textbf{Chest X-Ray} 
Wang \etal firstly created a multi-label large-scale Chest X-ray (CXR) diagnosis dataset by extracting 14 disease labels from clinical notes~\cite{Wang_2017_CVPR}.
Following works either using different backbones~\cite{Wang_2018_CVPR} or leveraging semantic disease relationship
to improve the CXR diagnosis performance~\cite{Zhang_Wang_Xu_Yu_Yuille_Xu_2020}. Although these works show ideal performance with large training dataset, the performance severely decreases if trained on a small CXR dataset, even on the disease with fewer training samples.
Collecting a large-scale and balanced CXR dataset is almost impossible in real-world clinical settings due to the expensive expert labelling cost and heterogeneous disease distribution. Alternatively, we aim at training a CXR classification model using smaller dataset by leveraging the radiologists' gaze information since the radiologist can learn to diagnose CXR with small size training samples.

\section{Our Proposed Framework}
\begin{figure}[t]
    \centering
    \includegraphics[width=0.95\linewidth]{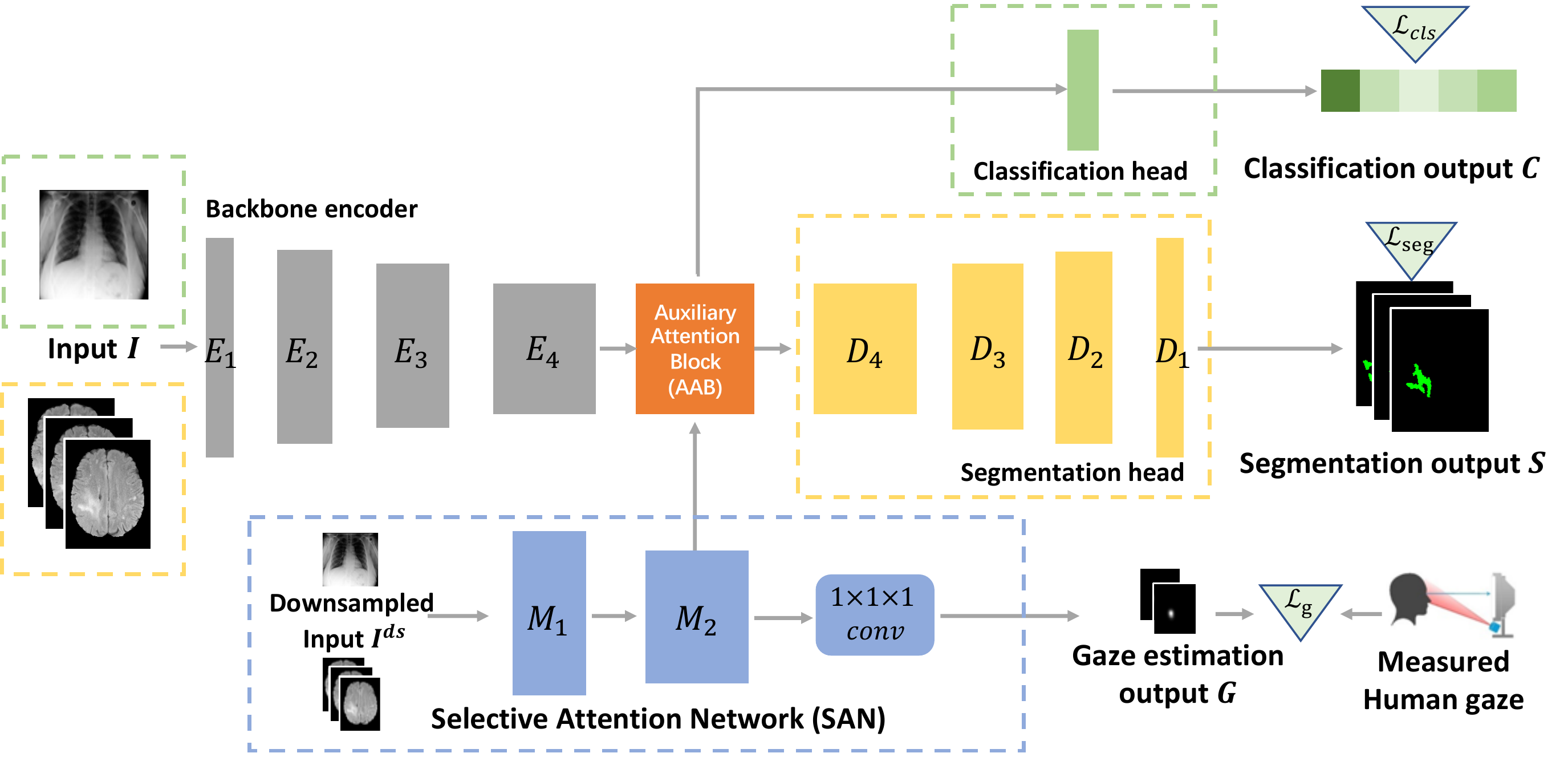}
    \caption{Overview of the proposed framework.}\vspace{-0.3cm}
    \label{fig:model}
\end{figure}
Here, we propose a framework that leverages gaze data as an auxiliary supervisory signal to improve the performance of different tasks, particularly under limited training data.
As illustrated in Fig~\ref{fig:model}, our proposed framework is composed of three parts: (1) a backbone encoder that encodes the input images into features, and different heads for performing corresponding tasks such as segmentation and classification. (2) A Selective Attention Network (SAN) that simulates the human selective attention mechanism and maintains a representation of overall cognition, and (3) in between the backbone encoder and the task-related heads, we insert a novel Auxiliary Attention Block (AAB) that leverages the cognition representation from SAN to guide the framework to attend to the disease-relevant clues.

\noindent\textbf{Backbone Encoder \& task-specific heads } 
By design, our framework can be applied to many existing medical imaging networks by integrating the SAN and AAB. Here, we demonstrate the 3D brain tumor segmentation task based on \cite{henry2020top} with 3D Unet as the backbone~\cite{3DUnet2016} and use a segmentation head. For the 2D chest X-ray classification, we select \cite{karargyris2021creation} with EfficientNet~\cite{tan2019efficientnet} as the backbone and use a classification head.

In the following part of this section, we showcase our framework on the task of 3D brain tumor segmentation~\cite{bakas2017advancing}. where the inputs are 3D scans of brain images $I \in \mathbb{R}^{D,W,H}$, where $D,W,H$ are the input depth, width, and height, respectively. 2D task of chest X-ray classification could be regarded as a special case of 3D with $D=1$, which also involves replacing 3D kernels with 2D ones. 

\subsection{Selective Attention Network}

SAN is inspired by the behavior of medical experts during screening, where they first selectively go through the whole slices of images quickly to acquire an overall cognition of the patient's condition.
As shown in the blue box of Fig~\ref{fig:model}, the Selective Attention Network (SAN) is a shallow network with 2 convolution groups that takes the $4\times$ down-sampled input images $I^{ds}$ as input. Each convolution block contains two $3\times3\times3$ convolution layers (with group normalization~\cite{wu2018group} and ReLU), followed by a 3D max-pooling layer. Notations such as $M_2$ denote the output of each convolution group.

When medical experts skim on the images, the selective attention mechanism would make them instinctively gaze at suspicious regions~\cite{Zhang2020GazeReview}. Therefore, we supervise the SAN with the collected medical expert's gaze. As shown in Fig~\ref{fig:model}, the output of the second convolution group $M_2$ is forwarded to a $1\times1\times1$ convolution layer to generate the gaze estimation map $G \in \mathbb{R}^{1,d,w,h}$, where $d, w, h$ are $\frac{1}{16}$ of the original image size. By learning to estimate human gaze, $M_2$ selectively encodes the information (\textit{e.g.} location, shape) of the suspicious regions. Therefore, we regard $M_2$ as a representation of selective attention and pass it to the backbone.

\subsection{Auxiliary Attention Block}
To better leverage the human underlying attention, we design a novel Auxiliary Attention Block and insert it between the encoder and the task heads. Specifically, our AAB fuses the backbone encoded feature $E_4$ with the latent representation of underlying attention $M_2$ from the SAN.

\begin{figure}[t]
    \centering
    \includegraphics[width=0.9\linewidth]{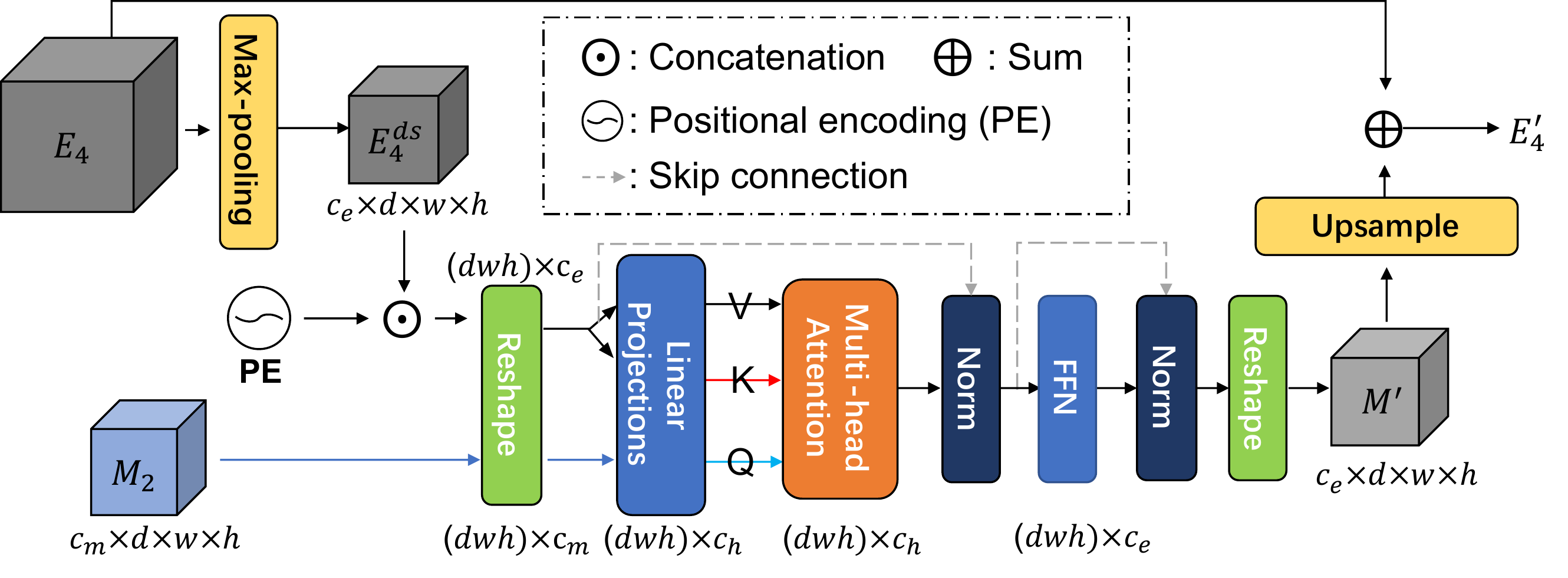}
    \caption{Architecture of the Auxiliary Attention Block. This block gets information from the SAN and feeds the backbone network information in the form of attention refined features.} \vspace{-0.2cm}
    \label{fig:gaze}
\end{figure}

We illustrate the proposed AAB in Fig.~\ref{fig:gaze}. The core of the AAB is a modified multi-head attention layer~\cite{vaswani2017attention}: using a \textit{query} vector generated from $M_2$, an attention weight on each location of the \textit{key} vector (generated from $E_4$) is computed. This layer thus generates attention weights to refine the feature by only focusing on the important, relevant parts. \revise{This is inspired from human visual attention mechanism, where human searches the visible space ($E_4$) using the selective attention ($M_2$) in mind~\cite{ungerleider2000mechanisms} as query.}
More formally speaking, the encoded feature $E_4$ is first down-sampled by a max pooling layer to match the spatial size of the cognition feature $M_2$, denoted as $E^{ds}_4 \in \mathbb{R}^{c_e,d,w,h}$. 
Following the previous success~\cite{nirkin2020hyperseg}, we add a fixed 3D positional encoding $P^{D,W,H} \in \mathbb{R}^{3,D,W,H}$ where in each position, $P^{D,W,H}_{i,j,k} = (\frac{2i-D+1}{D-1}, \frac{2j-W+1}{W-1}, \frac{2k-H+1}{H-1})$ and concatenate with $E^{ds}_4$. We then reshape the features by flattening all the spatial dimensions, and project $E^{ds}_4$ as the value $V$ and key $K$, and project $M_2$ as the query vector $Q$. The shape of $Q,K,V$ are all $\mathbb{R}^{(dwh)\times c_h}$ where $c_h=64$ is the hidden dimension. The core component in AAB is a multi-head attention block~\cite{vaswani2017attention}, where the query-key-value vectors are computed in the following manner:
\begin{gather}
    Att(Q, K, V) = Softmax(\frac{QK^T}{\sqrt{c_h}})V \\
    MultiHead(Q, K, V) = Concat(Att_1, Att_2, \cdots, Att_k)W^{O}.
\end{gather}
Here $W^{O}$ is a learnable weight matrix. 

As shown in Figure~\ref{fig:gaze}, the output of the multi-head attention block further passes through normalization layers and a small two-layer feed forward network (FFN) to get the refined feature $M'$. After up-sampling, $M'$ is then added with the original input feature $E_4$, forming the output $E'_4$. 
Finally, the refined feature $E'_4$ is forwarded to the segmentation head to generate the segmentation output $S$.

\subsection{Loss Functions and Model Training}
All the components of our framework can be trained using one hybrid loss function. As for the segmentation, denoting the segmentation ground-truth as $S^{gt}$, we use the dice loss function~\cite{milletari2016v,drozdzal2016importance} based on the previous success of brain tumor segmentation:
\begin{equation}
    \mathcal{L}_{seg} = \frac{1}{N}\sum_{n=1}^{N}\frac{2 \times S_n \cap S^{gt}_n }{S^2_n + (S^{gt}_n)^2},
\end{equation}
where $N$ is the number of output channels. 

The task of gaze estimation is trained using binary cross entropy loss~\cite{huang2018predicting}:
\begin{equation}
    \mathcal{L}_{g} = -\frac{1}{P}\sum\nolimits_{i=1}^P\left(G^{gt}_i\cdot log(G_i) + (1-G^{gt}_i) \cdot log(1 - G_i)\right),
\end{equation}
where $P$ is the total volume, and $G^{gt}$ is the measured expert's gaze positions obtained as explained in Section 4.1. 

The joint loss function is a weighted sum of the loss functions: $\mathcal{L} = w_1\mathcal{L}_{seg} + w_2\mathcal{L}_{g}$. For the task of chest X-ray classification, we apply the typical cross-entropy loss to replace $\mathcal{L}_{seg}$. 

\section{Experiment}
\vspace{-0.2cm}
In this section, we validate our proposed framework on two datasets: (1) brain tumor segmentation on the BraTS 2020 dataset~\cite{menze2014multimodal,bakas2018identifying,bakas2017advancing} and (2) disease classification on the MIMIC-CXR-gaze dataset~\cite{karargyris2021creation}.

\subsection{Brain tumor segmentation task}
\vspace{-0.1cm}
We conduct our experiments on the training set and validation set of the BraTS 2020 dataset. The dataset contains 3D MRI data of four modalities (T1, T1ce, T2, and FLAIR), each with shape $240\times 240\times 155$. This dataset also provides the survival state (short, medium, and long) of each patient. The size of the training set and validation set is 369 and 125, respectively. The targets of segmentation are three regions namely Enhancing Tumor (ET), Whole Tumor (WT), and Tumor Core (TC). The performance of 5-fold cross-validation on the training set and the performance on the validation set assessed by the online evaluation server are both used to evaluate our method. We report the dice score as the evaluation metric.

\begin{figure}[t]
    \centering
    \includegraphics[width=0.9\linewidth]{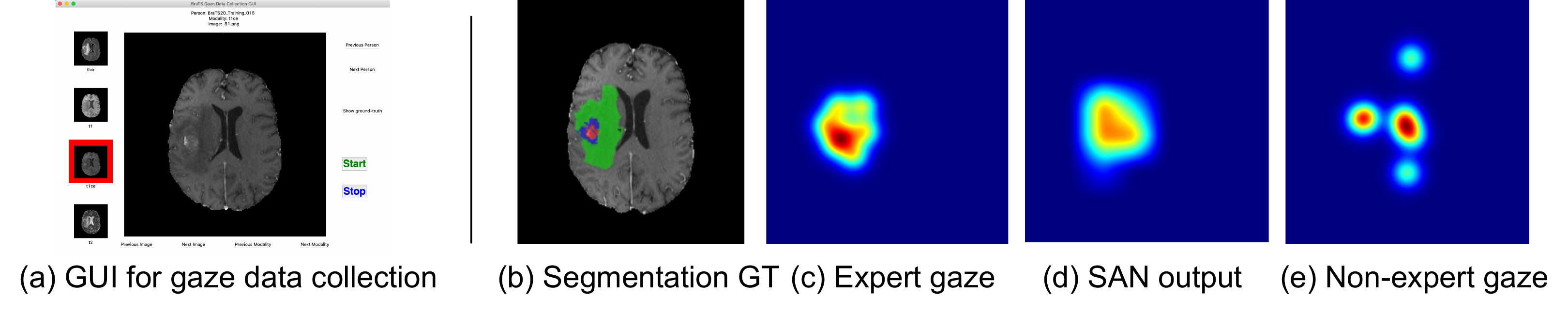}\vspace{-0.3cm}
    \caption{(a) GUI for Gaze data collection, (b) ground-truth segmentation labels, (c) gaze map of medical expert, (d) estimated gaze generated by SAN, (e) gaze map of non-expert.}\vspace{-0.3cm}
    \label{fig:data}
\end{figure}

\vspace{0.1cm}
\noindent\textbf{Gaze data collection}
Since BraTS dataset does not provide human gaze data, we collect gaze data on its training set using an eyeTribe eye-tracker\footnote{\url{https://theeyetribe.com/}}. We also designed a GUI as shown in Fig.~\ref{fig:data} to facilitate the gaze collection. This GUI is placed in a fixed position of the screen while the timestamps of the user's actions are automatically recorded. All the buttons in the GUI have corresponding shortcut keys to avoid looking at them during data collection.
During data collection, we ask the user to diagnose the severeness of each patient and find the tumor regions from the MRI images in the GUI similar to the daily screening.

Among all 369 data samples in the BraTS training set, we collect a total of 180 samples with gaze data from two-board certificated radiologists. \revise{The two doctors have 4 years and 2 years of professional experience, respectively. On average, each data sample is checked by the doctor for 38 seconds, and we use the code from~\cite{huang2018predicting} to filter out the saccade points and keep only the fixation points.} For comparison, we also collect the gaze data of these samples from two non-experts, however, they are not used in the training. The ground truth \revise{for} gaze estimation $G^{gt}$ is generated by adding a gaussian on the fixation points with kernel size 10, following the previous work~\cite{huang2018predicting}. An example is shown in Fig.~\ref{fig:data}, where (c) is the processed gaze map of an expert, and (e) is that of a non-expert. Note that our framework does not require human gaze data during inference.

\vspace{0.1cm}
\noindent\textbf{Model training} We empirically set $w_1=1, w_2=0.5$ in all of the experiments. The model is trained for 210 epochs using Ranger as the optimizer~\cite{yong2020gradient,zhang2019lookahead}, with an initial learning rate of 1e-4, and reduced by a cosine decay after 100 epochs. Including data augmentation, we keep other training and inference details the same as \textit{pipeline A} in \cite{henry2020top}, except that we do not use test time augmentation during inference. 

\vspace{0.1cm}
\noindent\textbf{Experiments }
We first show the experiment of 3D tumor segmentation. As for baseline, we use the backbone encoder alone (\textbf{Backbone}) without adding our framework. Here we mainly choose to use the backbone of one of the top solutions~\cite{henry2020top} from the BraTS 2020 challenge, since we experimentally verify that this backbone works better on small data compared with the state of the art~\cite{isensee2021nnu}. \revise{We also show that our proposed method can well cooperate with multiple backbones in the later part of this section.} The baseline \textbf{ours} indicates the backbone is equipped with the proposed Auxiliary Attention Block and the Selective Attention Network. To better evaluate the effectiveness of the proposed framework, since there is no previous work that uses gaze to enhance 3D segmentation task to the best of our knowledge, we construct another baseline, in which we add a new segmentation head with 1 channel output for gaze estimation, and use the estimated gaze as soft attention map during inference~\cite{li2021eye}. We denote this baseline as \textbf{+gaze$^{*}$}. \revise{To validate the effectiveness of the proposed AAB, we use a baseline \textbf{+SAN$^{*}$}, indicating that the information of SAN is not processed by AAB, but is fused to the backbone network by direct concatenation. To validate the improvement brought by gaze data, we use \textbf{ours-gaze} baseline, indicating that the SAN is trained also with ground-truth segmentation mask, but not gaze data.}

\renewcommand{\arraystretch}{1.1}
\begin{table}[t]
\centering
\scalebox{0.6}{
\begin{tabular}{@{}ccccc:cccc:cccc@{}}
\toprule
\multicolumn{1}{c}{Mean (std)} & \multicolumn{4}{c}{Enhancing Tumor}                              & \multicolumn{4}{c}{Whole Tumor} & \multicolumn{4}{c}{Tumor Core} \\ \cmidrule(lr){2-5}
\cmidrule(lr){6-9}
\cmidrule(lr){10-13}
Data ratio                & \multicolumn{1}{c}{0.2} & 0.3                  & 0.5 & 0.7 & 0.2  & 0.3  & 0.5  & 0.7 & 0.2  & 0.3  & 0.5  & 0.7  \\ \midrule
\multirow{2}{*}{Backbone}          & \multicolumn{1}{c}{
72.79}  & 74.57 & 76.32 & 77.78 & 85.41 & 85.70 & 85.82 & 88.82 & 80.11 & 81.20 & 82.45 & 84.24 \\
\vspace{0.08cm}
 &(0.022)&\multicolumn{1}{c}{(0.019)} & (0.027)& (0.007)& (0.018) & (0.016) & (0.022) & (0.006) & (0.017) & (0.017) & (0.012) & (0.016) \\ 
\multirow{2}{*}{+ gaze$^{*}$} & \multicolumn{1}{c}{72.97} & 74.24 & 75.77 & 78.01 & 83.42 & 85.22 & 86.01 & 88.71 & 79.97 & 81.14 & 82.03 & 84.65 \\ \vspace{0.08cm}
  & (0.015) & \multicolumn{1}{c}{(0.019)} & (0.014) & (0.024) & (0.038) & (0.023) & (0.016) & (0.020) & (0.020) & (0.009) & (0.015) & (0.012) \\ 
\multirow{2}{*}{\revise{+SAN$^{*}$}} & \multicolumn{1}{c}{71.84} & 72.06 & 74.37 & 75.90 & 82.21 & 84.58 & 85.10 & 86.56 & 77.74 & 79.66 & 80.73 & 82.34 \\ \vspace{0.08cm}
  & (0.025) & \multicolumn{1}{c}{(0.017)} & (0.018) & (0.030) & (0.035) & (0.027) & (0.024) & (0.023) & (0.016) & (0.013) & (0.018) & (0.015) \\ 
\multirow{2}{*}{\revise{ours-gaze}}   & \multicolumn{1}{c}{72.07} &  73.84 & 76.50 & 78.02 & 84.37 & 85.28 & 86.70 & 87.67 & 79.83 & 81.11 & 82.50 & 84.35 \\
\vspace{0.08cm}
 &  (0.020) & \multicolumn{1}{c}{(0.016)} & (0.017) & (0.023) & (0.021) & (0.028) & (0.022) & (0.015) & (0.017) & (0.010) & (0.018) & (0.014) \\ 
\multirow{2}{*}{ours}   & \multicolumn{1}{c}{74.69} &  76.46 & 79.05 & \textbf{81.43} & 85.05 & 85.44 & 89.14 & 89.13 & 82.51 & 82.27 & 85.46 & 85.42 \\
\vspace{0.08cm}
 &  (0.022) & \multicolumn{1}{c}{(0.007)} & (0.025) & (0.021) & (0.019) & (0.024) & (0.013) & (0.010) & (0.009) & (0.008) & (0.012) & (0.020) \\ 
\multirow{2}{*}{ours+surv}        & \multicolumn{1}{c}{\textbf{76.69}}  & \textbf{77.72} & \textbf{79.10} & 80.36 & \textbf{86.76} & \textbf{87.29} & \textbf{89.27} & \textbf{90.29} & \textbf{82.88} & \textbf{84.07} & \textbf{85.67} & \textbf{86.83}    \\
 &  (0.007) & \multicolumn{1}{c}{(0.012)} & (0.009) & (0.028) & (0.013) & (0.022) & (0.009) & (0.010) & (0.005) & (0.009) & (0.010) & (0.011)     \\ \bottomrule
\end{tabular}
}
\vspace{0.1cm}
\caption{Training set cross validation result (Dice score \%) using part of training data. The numbers in parenthesis are the std values.}\label{tab1}
\vspace{-0.2cm}
\end{table}

\begin{figure}[]
    \centering
    \includegraphics[width=0.8\linewidth]{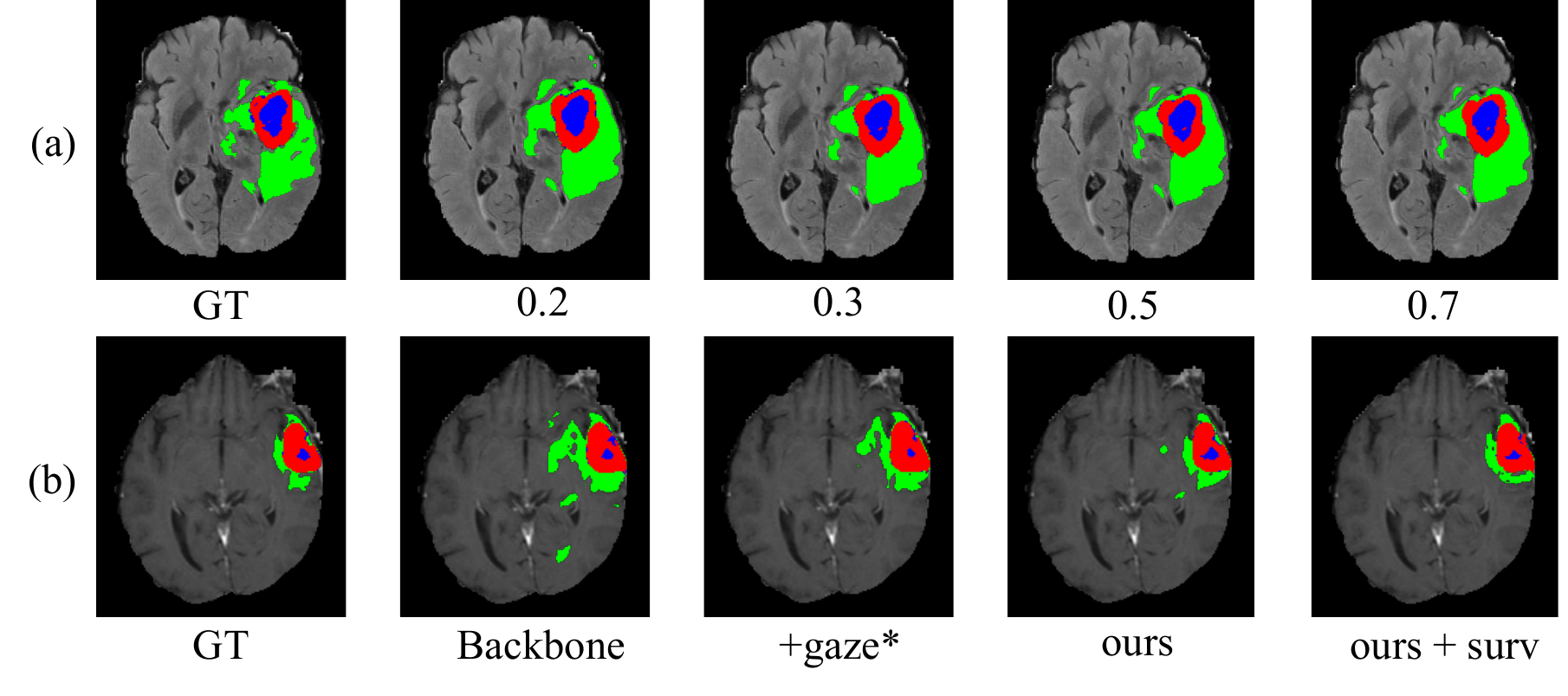} \vspace{-0.3cm}
    \caption{Qualitative comparison of different methods. (a) Performance of Backbone with our framework and multi-task learning (ours+surv), when using different amount of training data. (b) Performance of the backbone equipped with SAN and AAB (ours), and further multi-task training (ours+surv), using 20\% of training data.}
    \label{fig:quali}\vspace{-0.3cm}
\end{figure}

We also tested whether the framework can benefit from multi-task learning, by adding a classification head for survival state prediction, and denote this as \textbf{ours+surv}. To evaluate the effectiveness of our proposed framework using a smaller amount of training data, we conduct experiments using a part of training data that have collected gaze data under different proportions: 0.2, 0.3, 0.5, and 0.7. The experiment using a full set of training data, \revise{and the experiment using only the subset with gaze data,} can be found in the supplementary materials.

From Table~\ref{tab1}, the performance increase of \textbf{ours} compared with the backbone clearly demonstrates the advantage of our proposed framework when using small training data. From the comparison of \textbf{+gaze$^{*}$} and \textbf{ours} baselines, we can see that our proposed SAN and AAB can better leverage the auxiliary information provided by human gaze. \revise{The fact that \textbf{ours} outperforms \textbf{ours-gaze} } also demonstrates the human expert's gaze can provide complementary information other than segmentation masks, for example, experts' gaze would give more focus on the boundaries. \revise{The effectiveness of the proposed AAB is validated by the comparison between \textbf{ours} and \textbf{+SAN$^{*}$}}. Results on the validation set are in the supplementary materials, showing similar performances.

A qualitative example of the segmentation result is shown in Fig.~\ref{fig:quali}. The first row shows the performance of the \textbf{ours+surv} framework, when trained using different proportions of training data. The second row shows a comparison of different methods when using 20\% training data. From both rows, we can observe performance increase from left to right, which validates the benefit of our proposed framework. Fig.~\ref{fig:data} (d) shows the gaze estimation output from the SAN, which is highly consistent with the expert's gaze in Fig.~\ref{fig:data} (c). 

\begin{table}[]
\centering
\scalebox{0.8}{
\begin{tabular}{@{}ccccccc@{}}
\toprule
\multicolumn{1}{c}{Mean (std)} & \multicolumn{2}{c}{Enhancing Tumor}                              & \multicolumn{2}{c}{Whole Tumor} & \multicolumn{2}{c}{Tumor Core} \\ \cmidrule(lr){2-3}
\cmidrule(lr){4-5}
\cmidrule(lr){6-7}
Data ratio                & \multicolumn{1}{c}{0.2} & 0.5  & 0.2  & 0.5  & 0.2   & 0.5   \\ \midrule
EG & \textbf{74.69}(0.022) & \textbf{79.05}(0.025) & \textbf{85.05}(0.019) & \textbf{89.14}(0.013) & \textbf{82.51}(0.009) & \textbf{85.46}(0.012)  \\ 
n-EG  & 72.65(0.020) & 76.02(0.011) & 84.77(0.013) & 84.96(0.014) & 78.65(0.019) & 81.13(0.006)  \\ \bottomrule
\end{tabular}
}
\vspace{0.1cm}
\caption{Cross validation results on training set using expert gaze (EG) or non-expert gaze (n-EG) during training (Dice score \%). The numbers in parenthesis are the std values.}\label{tab3}\vspace{-0.3cm}
\end{table}

To further test the effect of gaze, we compare the performance of the network trained using expert gaze (EG) and non-expert gaze (n-EG) and show the results in Table~\ref{tab3}. The performance of the framework using expert gaze consistently outperforms the one using non-expert gaze. As shown in Fig.~\ref{fig:data}(e), non-expert gaze is more of a distraction for our framework as reading MRI data requires professional knowledge. 

\begin{table}[]
\centering
\scalebox{0.7}{
\begin{tabular}{@{}ccccccccccccc@{}}
\toprule
Type        & \multicolumn{4}{c}{ET}                              & \multicolumn{4}{c}{WT}                                                                      & \multicolumn{4}{c}{TC}                                                                                                              \\ \midrule
Data ratio  & \multicolumn{2}{c}{0.2}  & \multicolumn{2}{c}{0.5}  & \multicolumn{2}{c}{0.2}  & \multicolumn{2}{c}{0.5}                                & \multicolumn{2}{c}{0.2}                                          & \multicolumn{2}{c}{0.5}                                          \\ \cmidrule(l){2-13} 
    Metric    & Dice & H-95 & Dice & H-95 & Dice & H-95 & \multicolumn{1}{c}{Dice} & \multicolumn{1}{c}{H-95} & \multicolumn{1}{c}{Dice} & \multicolumn{1}{c}{H-95} & \multicolumn{1}{c}{Dice} & \multicolumn{1}{c}{H-95} \\ \midrule
Unet~\cite{henry2020top}      &  72.79 & 34.05 & 76.32 & 30.02 & 85.41 & 10.98 & 85.82  & 9.52 & 80.11 & 10.51 & 82.45 & 7.78  \\
nnUnet~\cite{isensee2021nnu}  & 71.86  & 38.52 & 73.74 & 28.42 & 83.79 & 13.53 & 85.67 & 9.90 & 79.87 & 17.07 & 80.78 & 16.21 \\
DMFNet~\cite{chen2019dmfnet} & 70.85 & 36.02 & 72.01 & 33.81 & 82.67 & 16.04 & 85.12 & 10.36 & 76.89 & 16.96  & 80.52 & 9.08 \\ \midrule
Unet~\cite{henry2020top}+ours      &  \textbf{74.69} & \textbf{33.61} & \textbf{79.05} & \textbf{21.28} & 85.05 & \textbf{10.61} & \textbf{89.14}  & 7.54 & \textbf{82.51} & \textbf{6.60} & \textbf{85.46} & 8.21  \\
nnUnet~\cite{isensee2021nnu}+ours  & 72.80  & 35.77 & 77.50 & 26.65 & \textbf{86.55} & 15.11 & 88.10 & 7.12 & 81.39 & 12.47 & 82.85 & 9.72 \\
DMFNet~\cite{chen2019dmfnet}+ours & 73.20 & 34.03 & 74.99 & 33.42 & 86.12 & 10.89 & 88.67 & \textbf{7.07} & 80.07 & 10.51  & 83.38 & \textbf{7.94} \\ \midrule
Unet+CBAM~\cite{woo2018cbam}   & 69.96  & 44.19 &  78.12  & 28.83 & 83.87 & 12.53 &  86.91 & 10.82 &  77.34  & 25.44 & 85.03  & 9.24 \\
Unet+TASN~\cite{zheng2019looking} & 70.03 & 44.20 & 73.70 & 36.19 & 83.61 & 17.07 & 83.99 & 13.53 & 79.81 & 11.68 & 81.96  & 17.46 \\ \bottomrule
\end{tabular}
}
\caption{\revise{Model cooperation with different backbones (top) and different attention modules (bottom). We experiment with Unet from \cite{henry2020top}, nnUnet~\cite{isensee2021nnu} and DMFNet~\cite{chen2019dmfnet} as backbones. We also replace our proposed framework with existing attention modules CBAM~\cite{woo2018cbam} and TASN~\cite{zheng2019looking} to test the effectiveness. Dice score (Dice) and the mean of Hausdorff95 distance (H-95) are shown as evaluation metrics.}}\label{tab4}\vspace{-0.3cm}
\end{table}
\revise{
It is worth noting that our method can work with different backbone models and improve their performance using gaze information as demonstrated in our experiment. We test Unet from \cite{henry2020top}, nnUnet~\cite{isensee2021nnu} and DMFNet~\cite{chen2019dmfnet}. nnUnet is the current state-of-the-art method for brain tumor segmentation, while DMFNet is an efficient backbone that could work real-time. As 
shown in Table~\ref{tab4}, with the help of our proposed modules, the performance of all backbones are increased. }

\revise{
For further testing the efficacy of the proposed method, we replace our method with several current common spatial attention methods. We choose CBAM~\cite{woo2018cbam} and TASN~\cite{zheng2019looking} for comparison, and supervise the attention map with ground truth gaze map. For TASN we supervise both the "structure attention" and "detailed attention". The results can be found in the last two rows of Table~\ref{tab4}. Clearly, we can see that directly apply existing spatial attention methods cannot help the backbone model under most circumstances. 
}

\subsection{Chest disease classification task}
We also conduct experiments on the MIMIC-CXR-gaze dataset~\cite{karargyris2021creation} for normal versus abnormal chest X-ray classification. This dataset contains 1083 CXR images with gaze collected from an American Board of Radiology certified radiologist. We use the pre-analyzed fixations to generate ground-truth of gaze map by putting an isotropic gaussian on each fixation point. Following \cite{irvin2019chexpert}, we use AUROC scores to evaluate the model performance on the classification of Atelectasis, Cardiomegaly, Consolidation, Edema, and Pleural Effusion. Other details including dataset split and model training can be found in the supplementary material. Note that the training is not conducted on the full MIMIC-CXR dataset~\cite{johnson2019mimic}.

In this experiment, following~\cite{karargyris2021creation} we use the encoder part of EfficientNet-b0~\cite{tan2019efficientnet} as the backbone encoder. Our experiments show that this backbone performs better than ResNet~\cite{he2016deep} and MobileNet~\cite{howard2017mobilenets} on small training data. The classification head output after sigmoid activation contains 5 channels, indicating the existence of each disease. We add two baselines for comparison. (1) Similarly to the brain tumor segmentation task, we add the \textbf{+gaze$^{*}$} baseline that directly adds a gaze prediction head on top of the backbone encoder, without using the proposed SAN and AAB. (2) We modify the recent classification work with gaze enhancement~\cite{liu2021goal} and use its gaze loss for supervising the gaze estimation head. 

\begin{table}[]
\centering
\scalebox{0.75}{
\begin{tabular}{@{}cccccccc@{}}
\toprule
Ratio                & Method & Average   & Atelectasis & Cardiomegaly & Consolidation & Edema & Pleural Effusion  \\ \midrule
\multirow{3}{*}{0.7} & Backbone &  55.14  &  61.06      &  57.72       &  32.56        & 62.83 &  61.55         \\
                     & +gaze$^{*}$ &  55.07    &    59.99    &     52.52    &     34.65     & 67.88 &     60.31     \\
                     & +gaze loss~\cite{liu2021goal} & 55.30    &    59.25    &     55.22    &     33.67     & 69.14 &     59.21       \\ 
                     & ours & \textbf{59.26}    &  65.68      &   61.31      &    33.23      & 74.14 &    61.94        \\ \midrule
\multirow{3}{*}{0.5} & Backbone & 50.43  &   52.30     &    54.47     &     28.06     & 66.40 &     50.94       \\
                     & +gaze$^{*}$ & 54.67    &    52.24    &     55.58    &     34.92     & 67.66 &     62.95       \\
                     & +gaze loss~\cite{liu2021goal} & 54.05    &    53.97    &     55.24    &     34.54     & 64.38 &     62.11       \\ 
                     & ours & \textbf{55.52}    &   61.85     &    55.86     &     31.35     & 67.70 &    60.86        \\ \midrule
\multirow{3}{*}{0.3} & Backbone & 46.60 &    42.02    &     49.77    &      36.32    & 59.87 &      46.01       \\
                     & +gaze$^{*}$ & 49.64    &    41.19    &     48.75    &     32.20     & 60.98 &     65.06       \\
                     & +gaze loss~\cite{liu2021goal} & 51.43    &    48.37    &     48.77    &     33.25     & 63.98 &     62.70       \\ 
                     & ours & \textbf{55.23}     &    56.92    &      53.63   &      32.40    & 69.10 &      64.10      \\ \midrule
\multirow{3}{*}{0.2}   & Backbone & 42.59 &     38.61 &     48.09    &     37.88     & 41.92 &       46.43      \\
                     & +gaze$^{*}$ & 43.96    &    42.22    &     45.80    &     36.08     & 41.75 &     53.93       \\
                     & +gaze loss~\cite{liu2021goal} & 45.68    &    46.51    &     46.70    &     35.64     & 47.16 &     52.38       \\ 
                     & ours & \textbf{48.11}    &    52.71    &     48.25    &     29.86     & 53.15 &     56.59       \\
\bottomrule
\end{tabular}
}
\vspace{0.1cm}
\caption{Results on the MIMIC-CXR-gaze dataset. AUROC score in percentage is used as the evaluation metric. Results are the average of the best 30 epochs. }\label{tab:chest}\vspace{-0.3cm}
\end{table}
Quantitative results are shown in Table~\ref{tab:chest}. Despite "Consolidation", where the backbone results surprisingly performs better when the amount of training data decreases, for all other four classes and the averaged performance, using our framework with gaze information, the performance can be significantly better when using limited training data. The comparison between our proposed framework and two baselines using gaze strongly proves that the performance improvement comes from the design of our framework, but not just from the gaze data. However, we can also see that in situations where the amount of data is limited, using non-invasively collected gaze as additional training data can benefit the model learning. We believe this shows the potential of using human gaze data in real-world clinical tasks. 

We also test the performance of different backbones (EfficientNet~\cite{tan2019efficientnet}, MobileNet-v2~\cite{howard2017mobilenets}, ResNet-18~\cite{he2016deep}) in our framework. The table of quantitative results can be found in the supplementary material. Among the three backbones, EfficientNet and MobileNet-v2 perform better than ResNet-18, which is reasonable since the amount of training data is limited, and ResNet is more complicated and requires more data for training when compared with the other two backbones.

\section{Conclusion}
In this work, we explored using human gaze to enhance model performance, especially on limited training data. We propose a novel framework that could leverage human gaze information to enhance multiple medical image analysis tasks such as brain tumor segmentation and chest disease classification. 
On both the BraTS 2020 dataset and the MIMIC-CXR-gaze dataset, our method achieves superior performance using a limited number of training samples when compared to existing methods. We also collected the real human gaze of both medical experts and non-experts on the BraTS 2020 dataset. 
Our experiment shows the potential of gaze and eye-tracker for real-world medical applications.

\small{
\section*{Acknowledgement}
This work was supported by JST under grant Moonshot R\&D Grant Number JPMJMS2011, ACT-X Grant Number JPMJAX190D, JST AIP Acceleration Research Grant Number JPMJCR20U1 and JSPS KAKENHI Grant Number JP20H04205, Japan.
}

\bibliography{egbib}
\end{document}